\documentclass[10pt,twocolumn,letterpaper]{article}

\usepackage{wacv}
\usepackage{times}
\usepackage{epsfig}
\usepackage{graphicx}
\usepackage{amsmath,amssymb,bm} 
\usepackage{booktabs}
\usepackage{colortbl}
\usepackage[normalem]{ulem}
\useunder{\uline}{\ul}{}
\usepackage{tablefootnote}
\usepackage{comment}
\usepackage{xcolor}
\usepackage{wrapfig}
\usepackage{mathtools}
\usepackage{tabularx}
\usepackage{multirow}
\usepackage{tabularray}
\usepackage{supertabular}

\usepackage{array}
\usepackage{ragged2e}
\usepackage{graphicx}
\usepackage{tablefootnote}

\def\sem_l{\mathcal{L}_{\text{sem}}}
\def\rec_l{\mathcal{L}_{\text{rec}}}
\def\clip_l{\mathcal{L}_{\text{clip}}}
\def\ortho_l{\mathcal{L}_{\text{ortho}}}
\def\disc_l{\mathcal{L}_{\text{mc}}}
\def\ps_l{\mathcal{L}_{\text{ps}}}
\def\emb_mod{\textbf{E}_{\text{mod}}}

\def\wacvPaperID{1352} 

\wacvfinalcopy 

\ifwacvfinal
\usepackage[breaklinks=true,bookmarks=false]{hyperref}
\else
\usepackage[pagebackref=true,breaklinks=true,colorlinks,bookmarks=false]{hyperref}
\fi
\usepackage[capitalize]{cleveref}

\begin{document}

\title{Modality-Aware Representation Learning for Zero-shot Sketch-based Image Retrieval}

\author{Eunyi Lyou$^1$\qquad Doyeon Lee$^1$ \qquad Jooeun Kim$^1$ \qquad Joonseok Lee$^{1,2}$\thanks{Corresponding author}\\
$^1$Seoul National University\qquad $^{1,2}$Google Research\\
{\tt\small $\{$onlyou0416, kje980714, joonseok$\}$@snu.ac.kr, omocomo83@gmail.com}
}

\maketitle
\thispagestyle{empty}

\begin{abstract}
Zero-shot learning offers an efficient solution for a machine learning model to treat unseen categories, avoiding exhaustive data collection. Zero-shot Sketch-based Image Retrieval (ZS-SBIR) simulates real-world scenarios where it is hard and costly to collect paired sketch-photo samples. We propose a novel framework that indirectly aligns sketches and photos by contrasting them through texts, removing the necessity of access to sketch-photo pairs. With an explicit modality encoding learned from data, our approach disentangles modality-agnostic semantics from modality-specific information, bridging the modality gap and enabling effective cross-modal content retrieval within a joint latent space. From comprehensive experiments, we verify the efficacy of the proposed model on ZS-SBIR, and it can be also applied to generalized and fine-grained settings.
\end{abstract}
\vspace{-0.3cm}
\section{Introduction}
\label{sec:intro}

Sketch-based image retrieval (SBIR) is a cross-view retrieval task, retrieving relevant natural images given an abstract and ambiguous sketch of an object.
Typically, a machine learning model is trained to map a photo and its corresponding sketch closely in a common latent space, trained on a set of paired samples.

However, it is usually challenging and often infeasible to acquire paired samples of free-hand sketches and natural photos across all potential objects.
For this reason, SBIR task has naturally adopted zero-shot learning, which has emerged to classify samples belonging to categories never seen at training.
In Zero-Shot Sketch-based Image Retrieval (ZS-SBIR)~\cite{Shen_2018_CVPR}, a model is trained on some collection of paired sketch-photo samples, and at testing time, it is asked to retrieve photos with an object unknown at training given its sketch. 

Under this zero-shot setting, prior works have studied semantic transfer by utilizing external semantic knowledge, such as label embeddings or hierarchical graphs~\cite{Liu_2019_ICCV, Dutta_2019_CVPR, 9102940,ribeiro2023sketchananchor, CHAUDHURI2020104003, chaudhuri22, Ge_Wang_Qi_Sun_Xu_Liao_2023}.
They provide insights for strategic handling of zero-shot scenarios in the presence of external supplementary information.
Recently, CLIP-AT~\cite{Sain_2023_CVPR} proposes employing comprehensive visual and textual embeddings from CLIP~\cite{clip}, trained on large-scale vision-language data.
Transferring the rich visual-linguistic relationship learned from the large data, this approach significantly improves the retrieval performance.


However, adapting off-the-shelf multimodal models such as CLIP~\cite{clip} to the ZS-SBIR task is non-trivial.
Modality gap~\cite{liang2022mind, shi2023towards} is a known phenomenon that image and text embeddings in a common space still tend to be clearly separated, even though the model is trained to embed images and texts based on their semantics, not based on their modalities.
Ideally, we would like to learn to map instances based on both their semantics and forms (modalities), where these two are disentangled. Then,
a simple nearest neighbor search within the target modality space would retrieve relevant items regardless of their forms.


To bridge the modality gap and fully utilize a multimodal foundation model, we propose Modality-Aware encoders for Sketch-based Image Retrieval (MA-SBIR), which explicitly and separately learn both modality-agnostic semantics and modality-specific information.
At a high level, our model is trained to transform a modality space into another within a shared latent space, by explicitly learning modality-specific nuance, separated from semantics.

A key advantage of this design is that the model can be trained without paired sketch-photo examples; it can be trained on a set of sketches and another set of photos labeled with a common vocabulary, without necessarily having one-to-one relationships.
Instead of directly learning to locate photos and sketches, we adopt an indirect approach through the category annotation (with their names as text modality).
Our model consumes an image (either a sketch or a photo) and its associated text at a time, and their semantic embeddings are aligned as previous models like CLIP.
At the same time, the model learns to distinguish sketches and photos with a separate modality encoding.
In this way, the model learns to represent semantics and modality, image by image, without requiring sketch-photo pairs.


By design, we believe our indirect alignment is more effective than the conventional direct alignment for the categorical SBIR, where a photo is considered correct if it is in the same category with the queried sketch (\emph{e.g.}, correct if both belong to clothes, regardless of their specific type or color).
In addition to this, we apply our proposed approach on two other relevant settings as well.
First, we test on the Generalized Zero-shot SBIR (GZS-SBIR)~\cite{9102940, chaudhuri22, Ge_Wang_Qi_Sun_Xu_Liao_2023, ribeiro2023sketchananchor, Lin_2023_CVPR}, which is a more realistic setting that the test set contains both seen and unseen classes.
This is to better simulate a real situation where performance on both is vital due to greater prevalence of seen classes.
Second, we test on a fine-grained (or instance-wise) SBIR setting (FG-SBIR), where only the one-to-one mapped photo is considered correct for each query sketch.
As our model is not designed to directly learn sketch-photo alignment, we expect sub-optimal performance on this setting.
For this reason, we adapt our model with a few modifications (Sec.~\ref{sec:method:instance}).

We summarize our contributions as follows:
\begin{enumerate}
    \setlength{\itemsep}{0pt}
    \setlength{\parskip}{0pt}
    \item We propose a novel method to align a joint embedding space, disentangling semantics from modality-specific information.
    \item Our proposed model indirectly aligns sketches and photos, removing the necessity of paired sketch-photo examples for training.
    \item We verify that our proposed method achieves the state-of-the-art performance on diverse zero-shot sketch-based image retrieval tasks.
\end{enumerate}

\section{Related Work}
\label{sec:related}
Sketch-based Image Retrieval (SBIR) can be categorized into two based on retrieval granularity: Category-level and Instance-level (fine-grained SBIR).
Category-level SBIR is to retrieve photos of an object in the same category among the candidate images covering
multiple categories, based on a given sketch.
For instance, given an image of a cat, category-level SBIR considers images of any kind of cats correct.
In instance-level SBIR, on the other hand, a sketch query should yield an exactly matching instance from images, not just those belonging to the same category.
In the above example, the images with the same cat instance are considered correctly retrieved.

\vspace{0.1cm} \noindent
\textbf{Category-level SBIR.}
Recent SBIR works have focused on aligning the shared embedding space of sketches and images by employing triplet ranking methods~\cite{qian2016, lin19, collomosse19, 8809264}, re-ranking scenarios~\cite{HUANG2018537, wang19}, or efficient hashing optimization~\cite{li17, zhang18, lu21} for this.
Furthermore, some approaches~\cite{zhang18, longteng17} align the distributions of images and sketches to train a generative model, forcing sketch and image representations to preserve their shared semantics.

The zero-shot framework, pioneered by \cite{Shen_2018_CVPR}, is designed to facilitate the retrieval of categories that are unseen at training.
ZS-SBIR often utilizes auxiliary information as a means of guiding previously unseen images into a common semantic space, \emph{e.g.}, from 
predefined class labels~\cite{Liu_2019_ICCV}, hierarchical graphs~\cite{Dutta_2019_CVPR, 9102940, ribeiro2023sketchananchor, CHAUDHURI2020104003, chaudhuri22}, and textual embeddings from a vision-language joint common space~\cite{Sain_2023_CVPR}.
Adversarial training~\cite{9102940, Dutta_2019_CVPR, CHAUDHURI2020104003} also has been employed.
\cite{chaudhuri22} aligns features from the intermediate and concluding layers of dual backbones, and  \cite{9775617} employs cross-domain mix-up strategies.
In order to disentangle domain and semantic features, gradient reversal layers~\cite{Dey_2019_CVPR} and separate modeling for each encoding~\cite{Tian_Xu_Shen_Yang_Shen_2022}.
Modern research has tackled catastrophic forgetting at fine-tuning to preserve the accumulated knowledge from the pre-training via knowledge distillation~\cite{Liu_2019_ICCV, wang22, tian21}, backbone sharing~\cite{9591307}, prompt learning~\cite{Sain_2023_CVPR}, and test-time adaptation~\cite{Sain_2022_CVPR}.

However, zero-shot models often overfits towards unknown categories.
To address this, a more realistic Generalized Zero-Shot Sketch-Based Image Retrieval (GZS-SBIR)~\cite{Dutta_2019_CVPR} setting is proposed, mixing seen and unseen classes in the test set.
A few recent works~\cite{9102940, chaudhuri22, Ge_Wang_Qi_Sun_Xu_Liao_2023, ribeiro2023sketchananchor, Lin_2023_CVPR} have explored this direction.

\vspace{0.1cm} \noindent
\textbf{Instance-level SBIR.}
Unlike the categorical SBIR, fine-grained (FG) SBIR aims to retrieve the exact target item at the instance level given a sketch image.
FG-SBIR models focus on enhancing pixel representations by incorporating spatial modules to account for detailed spatial positions~\cite{8237854}, cross-interaction modules to calculate patch-level similarities~\cite{9766165, Lin_2023_CVPR}, and utilizing randomized patch shuffling techniques~\cite{9157668, Sain_2023_CVPR}.

Aligned with the zero-shot setting in SBIR, fine-grained task is also tackled in zero-shot settings~\cite{Pang_2019_CVPR}.
Recent studies address cross-category generalization by leveraging the well-aligned CLIP~\cite{clip} space~\cite{Sain_2023_CVPR}, introducing cross-modal attention and patch-level matching~\cite{Lin_2023_CVPR}, and knowledge distillation from additional unlabelled photos~\cite{Sain_2023_CVPR2}.

\section{Method}
\label{sec:method}

\subsection{Problem Formulation}
\label{sec:method:problem}

Let $\mathcal{X}$ be a paired set of $n$ samples of a sketch and its corresponding real photo, where each pair is annotated with a class.
The $i$-th sample is denoted by $(\textbf{S}^{(i)}, \textbf{P}^{(i)}, c^{(i)}) \in \mathcal{X}$, where $\textbf{S}^{(i)} \in \mathbb{R}^{M \times N \times 3}$ is a sketch image of size $M \times N$, $\textbf{P}^{(i)}  \in \mathbb{R}^{M' \times N' \times 3}$ is its corresponding ground truth photo of size $M' \times N'$.
($\mathbf{S}$ and $\mathbf{P}$ may have different size.)
$c^{(i)} \in \mathcal{C}$ is the ground truth category of the pair, where $\mathcal{C}$ is a set of all categories.
Each category $c \in \mathcal{C}$ is transformed into a textual caption, formatted as ‘a photo of a $c$’,  denoted by $T \in |\mathcal{V}|^L$, where $\mathcal{V}$ is the vocabulary set and $L$ is the maximum length of the sentence.

On zero-shot tasks, $\mathcal{C}$ is further split to the seen classes ($\mathcal{C}_\text{s}$) and unseen classes ($\mathcal{C}_\text{u}$), used for training and testing, respectively, where
$\mathcal{C}_\text{s} \cap \mathcal{C}_\text{u}=\varnothing$ and $\mathcal{C}=\mathcal{C}_\text{s} \cup \mathcal{C}_\text{u}$.
The sketches and photos are split into train and test sets according to their labels; that is,
$\mathcal{X}_\text{train} = \{(\mathbf{S}^{(i)}, \mathbf{P}^{(i)}, T^{(i)}) \, |\, c^{(i)} \in \mathcal{C_\text{s}} \}$, 
$\mathcal{X}_\text{test} = \{(\mathbf{S}^{(i)}, \mathbf{P}^{(i)}) \, |\, c^{(i)} \in\mathcal{C_\text{u}} \}$.

In Sketch-based Image Retrieval (SBIR) task, a query sketch image $\mathbf{S}$ is given, and the model aims to retrieve relevant images from the candidate photos, either within the same category (category-level) or the exactly matched image instance (instance-level or fine-grained).
For category-level SBIR, the photos are considered relevant with any sketch within the same category; that is, $\mathcal{X}_c = \{ (\mathbf{S}^{(i)}, \mathbf{P}^{(j)})\ |\ c^{(i)} = c^{(j)} = c \}$ is a set of correct pairs for a class $c \in \mathcal{C}$.
In fine-graind SBIR, there always exists exactly one matched photo for each sketch; that is, $\mathcal{X}_i = \{ (\mathbf{S}^{(i)}, \mathbf{P}^{(i)})\}$ for a sample datum $i$.
For Generalized ZS-SBIR, randomly sub-sampled examples from $\mathcal{X}_\text{train}$ are added to $\mathcal{X}_\text{test}$, where $\mathcal{X}_\text{train}$ remains unchanged.


\subsection{The Proposed Model}
\label{sec:method:category}

\begin{figure*}
    \vspace{-0.6cm}
    \includegraphics[width=\linewidth]{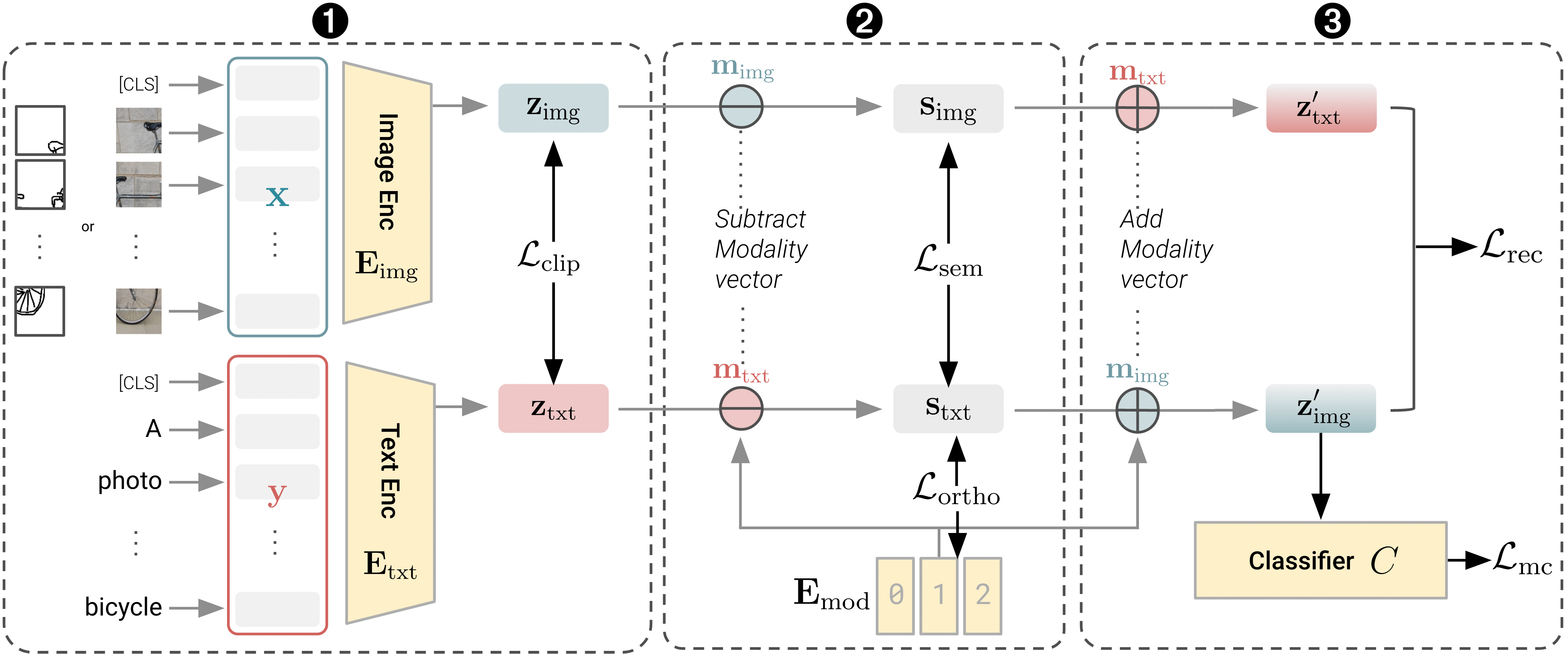}
    \caption{
        \textbf{Overview of our Architecture.} Given an image $\mathbf{x}$ (either a sketch or a photo) and a text $\mathbf{y}$, modality-specific encoders ($\mathbf{E}_{\text{img}}$ and $\mathbf{E}_{\text{txt}}$) embed them to $\mathbf{z}_{\text{img}}$ and $\mathbf{z}_{\text{txt}}$, respectively, where
        each $\mathbf{z}$ is a sum of semantic embedding $\mathbf{s}$ and modality encoding $\mathbf{m}$.
        We acquire semantic-only vectors ($\mathbf{s}_{\text{img}}$, $\mathbf{s}_{\text{txt}}$) from them by subtracting the modality encoding ($\mathbf{m}_{\text{img}}$, $\mathbf{m}_{\text{txt}}$).
        By adding the opposite modality encoding to the semantic embeddings, we reconstruct the image and text embeddings ($\mathbf{z}'_{\text{img}}$ and $\mathbf{z}'_{\text{txt}}$).
    }
    \label{fig_sbir}
    \vspace{-0.3cm}
\end{figure*}

\noindent
\textbf{Overview.}
The overall workflow of our approach for the category-level SBIR 
is illustrated in Fig.~\ref{fig_sbir}.
The first step comprises of modality-specific encoding of an input image, either a sketch $\mathbf{S}$ or a photo $\mathbf{P}$ (using the image encoder $\mathbf{E}_\text{img}$) and caption $T$ (using the text encoder $\mathbf{E}_\text{txt}$), and further alignment within a batch using CLIP loss, $\clip_l$.
In the second step, a learnable modality embedding vector is subtracted from the CLIP image ($\mathbf{z}_\text{img}$) and text ($\mathbf{z}_\text{txt}$) embeddings, to leave only their semantics, denoted by $\mathbf{s}_\text{img}$ and $\mathbf{s}_\text{txt}$, respectively.
We apply another semantic alignment loss, $\sem_l$, between them.
Lastly, the modality encoding is added back to the opposite modality, producing converted text ($\mathbf{z}'_\text{txt}$) and image ($\mathbf{z}'_\text{img}$) embeddings.
They are trained to reconstruct the original embedding ($\rec_l$) and to classify the target modality correctly ($\disc_l$).
This assists the model to disentangle common semantics from modality-specific information, eventually helping to transform one domain (\emph{e.g.}, sketch) to another (\emph{e.g.}, photo).
Once trained, the original and transformed embeddings are within the same embedding space, applicable for retrieval tasks.

\vspace{0.1cm} \noindent
\textbf{Inputs.}
The network receives a single image, either a sketch $\mathbf{S}$ or a photo $\mathbf{P}$, with a caption ($T$) as input.
Adopting the Vision Transformer~\cite{dosovitskiy2021an}, the input image is divided into multiple $P \times P$ patches $\in \mathbb{R}^{n_\text{patch}\times (P^2 \cdot 3)}$, where $n_\text{patch}$ is the resulting number of patches.
Each image patch is then linearly projected to a $d_\text{img}$-dimensional space.
Similarly, the caption is tokenized and vectorized, resulting in a sequence with $n_\text{seq}$ token embeddings of size $d_\text{txt}$.
An additional \texttt{[CLS]} token embedding is prepended to each sequence.
We denote the input image sequence by $\mathbf{x} \in \mathbb{R}^{(n_\text{patch}+1)\times d_\text{img}}$ and the text sequence by $\mathbf{y} \in \mathbb{R}^{(n_\text{seq}+1)\times d_\text{txt}}$.

\vspace{0.1cm} \noindent
\textbf{Indirect Alignment of Sketches and Photos.}
We use pre-trained CLIP~\cite{clip} image ($\textbf{E}_\text{img}$) and text encoders ($\textbf{E}_\text{txt}$), 
taking $\mathbf{x}$ and $\mathbf{y}$ as inputs, 
respectively.
From the output sequences, $\textbf{E}_\text{img}(\mathbf{x}) \in \mathbb{R}^{(n_\text{patch}+1)\times d_\text{img}}$ and $\textbf{E}_\text{txt}(\mathbf{y}) \in \mathbb{R}^{(n_\text{seq}+1)\times d_\text{txt}}$, we take the representations corresponding to $\texttt{[CLS]}$, and linearly map them to a common embedding size $d$.
We denote the resulting embeddings by $\mathbf{z}_\text{img}\in\mathbb{R}^d$ and $\mathbf{z}_\text{txt}\in\mathbb{R}^d$, respectively, and they are our base image and text representations.

Given $\mathbf{z}_\text{img}$ and $\mathbf{z}_\text{txt}$, we further align the embeddings by minimizing the following loss:
\begin{equation}
  \mathcal{L}_{\text{clip}} = \frac{1}{2}\left( \text{CLIP}(\mathbf{z}_\text{img}, \mathbf{z}_\text{txt})+\text{CLIP}(\mathbf{z}_\text{txt}, \mathbf{z}_\text{img}) \right), \label{clip-loss-sbir}
\end{equation}
\begin{equation}
  \text{where} \ \ \text{CLIP}(\mathbf{a}, \mathbf{b}) = 
  -\log\frac{\exp{(\mathbf{a} \cdot \mathbf{b}/\tau)}}{\sum^{B}_{j=1}{\exp{(\mathbf{a} \cdot \mathbf{b}_{j}/\tau)}}}.
  \label{clip-loss-def}
\end{equation}

Our approach is distinguished from existing methods in that ours does not have a direct mechanism to align sketches and photos within the image modality, while the subtle differences between them are indirectly aligned through texts.
This approach is opposed to the direct alignment via triplet learning in existing works~\cite{qian2016, lin19, collomosse19, 8809264}.
The biggest advantage of our approach over the previous direct triplet learning is that ours does not require paired training examples of sketches and photos, which are costly to collect.
In order to directly train the model to distinguish the two, previous models have relied on positive pairs of a sketch and a photo.
With our indirect approach, however, no explicit positive relationship between a sketch and a photo is utilized.
Instead, both sketches and photos are embedded via a common image encoder and aligned with the associated text, taking advantage of the intricate representation capacity of the CLIP.
This capability of leveraging unpaired datasets addresses the well-known issue of data scarcity in the community, aligning with the efforts of \cite{Bhunia_2021_CVPR, Sain_2023_CVPR2} to tackle limited data availability.
Our design also simplifies the contrastive loss term, removing the need for negative sampling.

\vspace{0.1cm} \noindent
\textbf{Modality Encoding.}
Since our model uses a common visual encoder for sketches and photos, we need an additional mechanism to distinguish them.
For this, we introduce modality encoder $\mathbf{E}_\text{mod}$, composed of learnable encoding ($\mathbf{m} \in \mathbb{R}^d$) for each modality. Specifically, we assign a unique index to each specific modality; for example, 
sketches, photos, and texts are assigned with 0, 1, and 2, respectively.
Under our design, the model \emph{learns} to represent the modality itself, regardless of its nature, in its modality encoding $\mathbf{m}$.

Under this setting, our model is trained to separate its latent space into its constituent semantic and modality components.
Once this is achieved, one can convert an embedding from one domain to another simply by
\begin{equation}
  \mathbf{x}_\text{trg} = \mathbf{x}_\text{src} - \mathbf{m}_\text{src} + \mathbf{m}_\text{trg},
\end{equation}
where $\mathbf{x} \in \mathbb{R}^d$ are the learned embedding (containing both semantics and modality), $\mathbf{m} \in \mathbb{R}^d$ are the modality encodings corresponding to each dataset, either source (src) and target (trg).
We expect the converted representation in this way to yield improved performance on the retrieval tasks.


\vspace{0.1cm} \noindent
\textbf{Visual-Text Alignment.}
From the CLIP embeddings, $\mathbf{z}_\text{img} \in \mathbb{R}^d$ and $\mathbf{z}_\text{txt} \in \mathbb{R}^d$, we subtract their modality encodings, $\mathbf{m}_\text{img}  \in \mathbb{R}^d$ and $\mathbf{m}_\text{txt}  \in \mathbb{R}^d$, respectively.
We get image and text embeddings purely based on their semantics:
\begin{align}
  \mathbf{s}_\text{img} = \text{N}(\mathbf{z}_\text{img} - \mathbf{m}_\text{img}), \ \mathbf{s}_\text{txt} = \text{N}(\mathbf{z}_\text{txt} - \mathbf{m}_\text{txt}),
\end{align}
where $\text{N}(x) = \mathbf{x} / \|\mathbf{x}\|_2$, indicating the normalization operator.
For a positive paired example of an image and its associated text, we train the model to keep their semantic representations similar.
Specifically, we consider the normalized average of image and text embeddings as its general semantic representations, denoted by $\mathbf{z}$, and both semantic representations are encouraged to be close to it.
Formally, we minimize the following semantic loss $\mathcal{L}_\text{sem}$:
\begin{equation}
  \mathcal{L}_\text{sem} =
  -\cos \left(\mathbf{s}_\text{img},
  \mathbf{z} \right)
  -\cos \left(\mathbf{s}_\text{txt},
  \mathbf{z} \right),
\end{equation}
where $\mathbf{z} = \text{N} ((\mathbf{z}_\text{img} + \mathbf{z}_\text{txt}) / 2)$.
We use cosine similarity, but other similarity or distance functions may be applicable.


\vspace{0.1cm} \noindent
\textbf{Cross-modal Reconstruction.}
On the purely semantic representations, $\mathbf{s}_\text{img}$ and $\mathbf{s}_\text{txt}$, we add the modality encoding from the opposite modality, producing reconstructed text and image embeddings, denoted by $\mathbf{z}'_\text{txt}$ and $\mathbf{z}'_\text{img}$:
\begin{equation}
  \mathbf{z}'_\text{txt} = \text{N}(\mathbf{s}_\text{img} + \mathbf{m}_\text{txt}), \  \mathbf{z}'_\text{img} = \text{N}(\mathbf{s}_\text{txt} + \mathbf{m}_\text{img}).
  \label{converted-vecter-formula}
\end{equation}
We apply the following loss to ensure the reconstructed embedding maintain similar directions to the original:
\begin{equation}
  \mathcal{L}_\text{rec} =
  -\cos \left( \mathbf{z}'_\text{txt}, \mathbf{z}_\text{txt} \right)
  -\cos \left( \mathbf{z}'_\text{img}, \mathbf{z}_\text{img} \right).
\end{equation}

In addition, to reconstruct more precisely, we introduce a modality classifier on the reconstructed embeddings, $\mathbf{z}'_\text{img}$ and $\mathbf{z}'_\text{txt}$.
For each, we minimize a CE loss $\mathcal{L}_\text{mc}$ defined as
\begin{align}
  \mathcal{L}_\text{mc} = 
  \sum_{j=1}^{C} m_{j} \log(\hat{m}_{j}),
\end{align}
where $\hat{m}_{j} \in \mathbb{R}^C$ is the predicted logits for the input embedding to belong to each modality, $m_j$ is the one-hot encoding of its ground-truth modality, and $C$ is the number of modalities.
By minimizing the classification loss, we ensure that the reconstructed $\mathbf{z}'_\text{\{txt,img\}}$ originate from separate classes recognizable by the classifier, irrespective of their initial modality.
This approach helps emphasize the distinctions between different modalities.

\vspace{0.1cm} \noindent
\textbf{Orthogonal Regularization.}
While the previously introduced losses ensure the disentanglement of semantic and modality information, this can be enhanced by imposing orthogonality between two directions.
Specifically, we design the orthogonality regularizer $\mathcal{L}_\text{ortho}$ as follows:
\begin{align}
  \mathcal{L}_\text{ortho} = \frac{1}{C} \sum_{j=1}^{C} | \textbf{z} \cdot \mathbf{m}_j |,
\end{align}
where $\mathbf{m}_j$ is the $j$'th modality embedding, and $\mathbf{z} = \text{N}(\mathbf{z}_\text{img} + \mathbf{z}_\text{txt} /2)$.
This leads to the alignment of the dot products between all vectors in semantic matrix $E_s$ and modality matrix $E_m$ towards zero, consequently enforcing orthogonality.
While a dot product of zero indicates either perpendicularity between non-zero vectors or one of the vectors being zero, we preemptively prevent the latter through other previously mentioned loss terms. For instance, the uniqueness of modality vectors is ensured by minimizing $\mathcal{L}_\text{mc}$, 
and semantic vectors resist convergence to zero due to the preservation of uniformity and alignment in the latent space, via $\mathcal{L}_{\text{clip}}$, as observed in \cite{pmlr-v119-wang20k}.

\vspace{0.1cm} \noindent
\textbf{Overall Training Objective.}
Our model is trained to minimize the following:
\begin{align}
  \mathcal{L} = 
  \clip_l
  + \lambda_{\text{sem}}\sem_l
  + \lambda_{\text{mc}} \mathcal{L}_\text{mc}
  + \lambda_{\text{rec}}\rec_l
  + \lambda_{\text{ortho}}\ortho_l, \nonumber
\end{align}
where all $\lambda$s are hyperparameters to control relative importance between each loss.
See Sec.~\ref{implementation-details} for more details.

\subsection{Extension to Instance-level SBIR}
\label{sec:method:instance}

\begin{figure}
    \centering
    \vspace{-0.4cm}
    \includegraphics[width=0.85\linewidth]{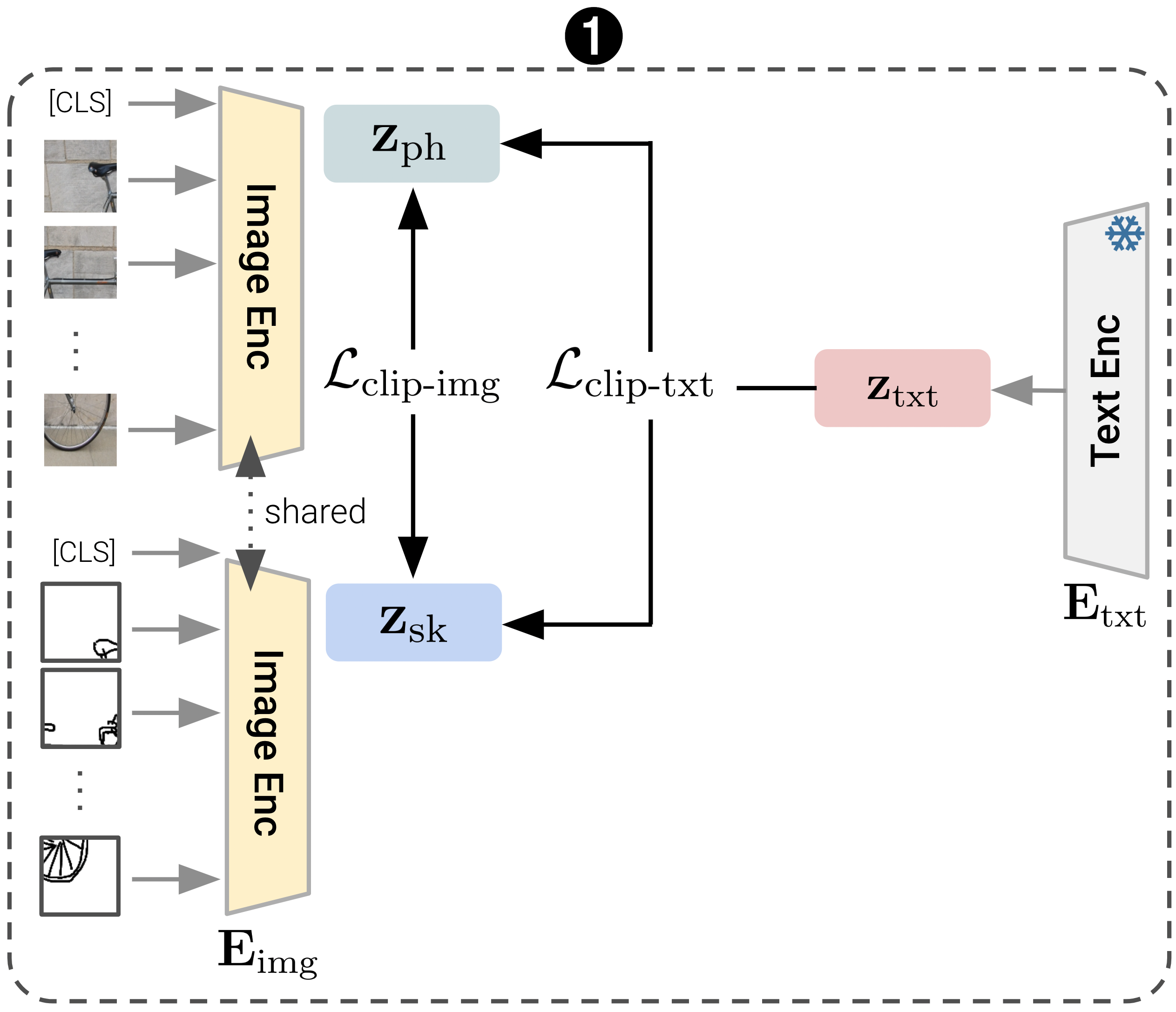}
    \caption{
        \textbf{Overview of FG-SBIR model architecture.} Unlike for categorical SBIR, fine-grained version takes paired sketches and photos as input to obtain latent vectors ($\mathbf{z}_\text{\{ph,sk\}}$), followed by step 2 and 3 described in \cref{fig_sbir}.
    }
    \label{fig_fg_sbir}
\end{figure}

Fine-grained SBIR (FG-SBIR) task aims to retrieve precisely matched photos from a sketch query. Unlike the category-level SBIR, merely retrieving images in the same category is insufficient.
Intuitively, our proposed approach in Sec.~\ref{sec:method:category} is not suitable for this task, as it mainly relies on the indirect connection between sketches and photos through coarse-grained textual descriptions, without direct enforcement based on paired examples.
For this reason, our vanilla model is not suitable for finer-grained alignment.

However, we pose a question: can our model still perform well, if it is trained on the paired data?
To answer this question, we adapt our model to the instance-level SBIR setting with some modifications guided by the inherent nature of the task.
Specifically, we take a photo-sketch-text triplet as input, instead of an image-text pair, although we still use a common image encoder.
It is inevitable in this fine-grained setting to take paired photo-sketch examples.
Some loss functions are also modified, explained in detail subsequently.
However, there is no major change to the model architecture, other than the straightforward extension to use three encoders depicted in Fig.~\ref{fig_fg_sbir}.

\vspace{0.1cm} \noindent
\textbf{Inputs and CLIP Loss.}
To learn more sophisticated semantic distribution in the sketch and photograph modalities, a training example is composed of a triplet $(\mathbf{S}, \mathbf{P}, T)$, where $\mathbf{S}$, $\mathbf{P}$, and $T$ indicate a sketch, its corresponding photo, and their textual caption, respectively.
Each of these three is considered a distinguished modality, assigned with its own modality index.
Subsequently, we extract modality-specific features:
$\mathbf{z}_\text{sk} = \textbf{E}_{\text{img}}(\mathbf{S}) \in \mathbb{R}^d$, $\mathbf{z}_\text{ph} = \textbf{E}_{\text{img}}(\mathbf{P}) \in \mathbb{R}^d$, and $\mathbf{z}_\text{txt} = \textbf{E}_{\text{txt}}(T) \in \mathbb{R}^d$, respectively, where $\mathbf{E}_\text{\{img,txt\}}$ are pre-trained modality-specific encoders.

The biggest change for fine-grained model is introduction of direct sketch-photo alignment.
Specifically, we add another contrastive loss designed for sketch-photo alignment, defined as follows:
\begin{equation}
\mathcal{L}_{\text{clip-img}} = \frac{1}{2}\left( \text{CLIP}(\mathbf{z}_\text{sk}, \mathbf{z}_\text{ph})+\text{CLIP}(\mathbf{z}_\text{ph}, \mathbf{z}_\text{sk}) \right),
\end{equation}
where CLIP$(\mathbf{a}, \mathbf{b})$ is defined in Eq.~\eqref{clip-loss-def}.We balance this direct alignment with the existing text-guided indirect alignment from Eq.~\eqref{clip-loss-sbir} by convex combination:
\begin{align}
  \clip_l = \lambda_{\text{clip-img}}\mathcal{L}_{\text{clip-img}} + (1-\lambda_{\text{clip-img}})\mathcal{L}_{\text{clip-txt}},
\end{align}
where $\mathcal{L}_{\text{clip-txt}}$ is equivalent as Eq.~\eqref{clip-loss-sbir} but we use different notation to ensure that notation $\clip_l$ continues to represent the overall contrastive loss.
With a hyperparameter $\lambda_\text{clip-img}$, we adjust the relative influence from the two types of contrastive losses.


\vspace{0.1cm} \noindent
\textbf{Additional Regularizations.}
Beyond comprehending the semantics of images, grasping the structural aspects, such as the position of an instance or edge detection within the image, is another crucial dimension in FG-SBIR.
To effectively handle geometric attributes, we adopt patch shuffling, aligning with prior studies~\cite{9766165, Lin_2023_CVPR, 9157668, Sain_2023_CVPR} with  similar objectives. 
Initially, non-overlapping patches from the sketch and the photo are randomly shuffled using a permutation function $\pi$, 
and then fed into the image encoder $\mathbf{E}_\text{img}$.
We take the representation corresponding to the \texttt{[CLS]} token as the sketch and photo embedding, denoted by $\mathbf{z}_\text{sk}$ and $\mathbf{z}_\text{ph}$, respectively.
We additionally minimize the following patch shuffling loss, allowing push-and-pull between these two:
\begin{align}
\mathcal{L}_{\text{ps}} &= \frac{1}{2}\left( \text{CLIP}(\mathbf{z}_\text{sk}, \mathbf{z}_\text{ph})+\text{CLIP}(\mathbf{z}_\text{ph}, \mathbf{z}_\text{sk}) \right).
\end{align}
\vspace{0.1cm} \noindent
\textbf{Overall Training Objective.}
With $\lambda$s as hyperparameters, the overall loss is given by
\begin{align}
  \mathcal{L} = \ &\clip_l +\lambda_{\text{ps}}\ps_l+\lambda_{\text{sem}}\sem_l \\
  &+ \lambda_{\text{mc}}\disc_l + \lambda_{\text{rec}}\rec_l + \lambda_{\text{ortho}}\ortho_l. \nonumber
\end{align}

\subsection{Implementation Details}
\label{implementation-details}
We set $d$ to 512 and 768 for ZS-SBIR and FG-ZS-SBIR, respectively. We use pretrained CLIP~\cite{clip} and ViT (ViT-B/32 for categorical and ViT-B/16 for fine-grained) as backbone. For FG-ZS-SBIR, $\textbf{z}_\text{img} \in \mathbb{R}^{768}$ is extracted from the $\mathbf{E}_\text{img}$ without the last projection layer. The text embedding is projected to the same size by a projection layer. Also, we strategically freeze the text encoder during training to enhance efficiency, supported by our findings that this maintains performance unaffected for FG-ZS-SBIR. Modality classifier is implemented with a FC layer. 

We use AdamW optimizer with early stopping.
The learning rate for the image/text encoder is set deliberately lower than the rest, to prevent catastrophic forgetting: (5e-8, 5e-3) for ZS-SBIR, and (5e-5, 1e-2) for FG-ZS-SBIR.
We use batch size of 256 on Sketch-Ext and TU-Berlin-Ext for ZS-SBIR, while 64 on QuickDraw-Ext for ZS-SBIR and on Sketchy for FG-ZS-SBIR.
We grid search the weight coefficients $\lambda_{\text{\{sem, mc, rec, ortho\}}}$ and set to (0.1, 0.1, 0.02, 0) for Sketchy-Ext, (0.1, 0.1, 0, 0) for TU-Berlin-Ext, and (0.2, 0.2, 0.2, 0.2) for QuickDraw-Ext.
For FG-ZS-SBIR, we use (0.5, 0.1, 0.02, 0.1, 0.02) for $\lambda_{\text{\{ps, sem, mc, rec, ortho\}}}$. We report mean performance of three independent experiments.
We implement our model with PyTorch~\cite{pytorch} and experiment on a NVIDIA RTX A6000 GPU.


\section{Experimental Settings}
\label{sec:experiments}

\noindent
\textbf{Datasets.}
We evaluate our method on well-known ZS-SBIR datasets: Sketchy Extended, TU-Berlin Extended and QuickDraw Extended. Sketchy Extended~\cite{Liu_2017_CVPR} consists of 73,002 images on 125 categories, on average 604 sketches and 584 images per class, extending Sketchy~\cite{10.1145/2897824.2925954}.
For zero-shot experiments, we set aside 21 classes that are not present in the 1,000 classes of ImageNet for testing, leaving the rest 104 classes for training, following \cite{Yelamarthi_2018_ECCV}.


TU-Berlin Extended~\cite{Liu_2017_CVPR} extends the TU-Berlin~\cite{eitz2012hdhso}, originally designed for sketch classification and composed of 20,000 sketches on 250 object categories in a balanced manner, by incorporating 204,489 natural images from \cite{7780494}. After this extension, each category has 787 images on average, but highly imbalanced.
We follow the partition protocol in \cite{Shen_2018_CVPR}, where 30 classes are randomly selected for testing, leaving the rest 220 classes for training. Due to the significant imbalances in the numbers of real images in each class, \cite{Shen_2018_CVPR} also made sure that each test category has at least 400 images when choosing the test set.

QuickDraw Extended~\cite{Dey_2019_CVPR} is a large-scale dataset designed for ZS-SBIR.
Using Google Quick, Draw! data, it contains 110 categories with 330,000 sketches, including 3,000 amateur sketches per category.
It also has 204,000 images taken from Flickr tagged with the corresponding label. For split, we follow \cite{Yelamarthi_2018_ECCV,Dey_2019_CVPR} to ensure the 30 test classes free from ImageNet, using the rest 80 classes for training.

\vspace{0.1cm} \noindent
\textbf{Baselines.}
We compare our method with state-of-the-art methods for ZS-SBIR and FG-ZS-SBIR.
For ZS-SBIR, we compare with CNN-based models (SEM-PCYC~\cite{Dutta_2019_CVPR}, SAKE~\cite{Liu_2019_ICCV}, OCEAN~\cite{9102940}, BDA~\cite{chaudhuri22}, and Sketch-3T~\cite{Sain_2022_CVPR}) and ViT-based models (TVT~\cite{Tian_Xu_Shen_Yang_Shen_2022}, PSKD~\cite{wang22}, SaA~\cite{ribeiro2023sketchananchor}, ZSE~\cite{Lin_2023_CVPR}, and CLIP-AT\cite{Sain_2023_CVPR}). For generalized ZS-SBIR, we additional compare with STL\cite{Ge_Wang_Qi_Sun_Xu_Liao_2023}.


\vspace{0.1cm} \noindent
\textbf{Our Model Variants.}
We present three variations of our proposed approach. Initially, we train our model solely with $\clip_l$ to assess the impact of indirect alignment technique, labeled with \textbf{[clip]}. On our full model, we may use the original embeddings $\mathbf{z}_\text{\{img, txt\}}$, marked with \textbf{[original]}, or the converted ones $\mathbf{z}'_\text{\{img, txt\}}$, labeled with \textbf{[converted]}.

\vspace{0.1cm} \noindent
\textbf{Evaluations Metrics.}
We use two standard retrieval metrics: mean Average Precision (mAP@$k$) and Precision (Prec@$k$). mAP calculates average precision at different recall levels, while precision measures relevance from the top $k$ retrieved items. $k$ is set following the standard in literature: $k = 200$ on Sketchy-Ext~\cite{Liu_2017_CVPR}, $k = 100$ on TU-Berlin-Ext~\cite{Liu_2017_CVPR} only for Precision, and $k = 200$ on QuickDraw-Ext~\cite{Dey_2019_CVPR} only for Precision.
For FG-SBIR, we report accuracy@\{1, 10\}, the ratio of sketches correctly matching with top-$k$ retrieved photographs, following previous studies.


\section{Results and Discussion}
\label{sec:result}

\subsection{Performance Analysis}

\renewcommand{\arraystretch}{0.7}
\begin{table}
\centering
\scriptsize
\setlength{\extrarowheight}{0pt}
\addtolength{\extrarowheight}{\aboverulesep}
\addtolength{\extrarowheight}{\belowrulesep}
\setlength{\aboverulesep}{0pt}
\setlength{\belowrulesep}{0pt}
\renewcommand{\tabcolsep}{3pt}
\caption{\centering Comparison for Categorical ZS-SBIR}

\begin{tabular}{clcccccc} 
\toprule
\multicolumn{2}{c}{\multirow{2}{*}{Model}}& \multicolumn{2}{c}{Sketchy-Ext}& \multicolumn{2}{c}{TU-Berlin-Ext}& \multicolumn{2}{c}{QuickDraw-Ext}\\ 
\cline{3-8}
\multicolumn{2}{c}{}& \multicolumn{1}{c}{m@200}& \multicolumn{1}{c}{P@200}& \multicolumn{1}{c}{m@all}& \multicolumn{1}{c}{P@100}& \multicolumn{1}{c}{m@all}& \multicolumn{1}{c}{P@200}\\ 
\hline
\multirow{5}{*}{CNN} & SEM-PCYC~\cite{Dutta_2019_CVPR}& -& -& 29.7& 42.6& -& -\\
                     & SAKE~\cite{Liu_2019_ICCV}& 49.7& 59.8& 47.5& 59.9& -& -\\
                     & OCEAN~\cite{9102940}& -& -& 33.3& 46.7& -& -\\
                     & BDA~\cite{chaudhuri22}& 45.8& 55.6& 37.4& 50.4& 15.4& 35.5\\
                     & Sketch-3T~\cite{Sain_2022_CVPR}& -& 62.4& 50.7& -& -& -\\ 
\hline
\multirow{9}{*}{ViT} & TVT~\cite{Tian_Xu_Shen_Yang_Shen_2022}& 53.1& 61.8& 48.4& 66.2& 14.9& 29.3\\
                     & PSKD~\cite{wang22}& 56.0& 64.5& 50.2& 66.2& 15.0& 29.8\\
                     & SaA~\cite{ribeiro2023sketchananchor}& 53.5& 63.0& 49.0& 60.8& 14.8& -\\
                     & ZSE[Ret]~\cite{Lin_2023_CVPR}& 50.4& 60.2& 56.9& 63.7& 14.2& 20.2\\
                     & ZSE[RN]~\cite{Lin_2023_CVPR}&52.5&62.4&54.2&65.7&14.5&21.6\\
                     & CLIP-AT$^{*}$~\cite{Sain_2023_CVPR}&63.6& 71.0& 65.9& 76.7& 29.3& 36.4\\
                     & {\cellcolor[rgb]{0.925,0.925,0.925}}Ours [clip] & {\cellcolor[rgb]{0.925,0.925,0.925}}\uline{68.5}   & {\cellcolor[rgb]{0.925,0.925,0.925}}\uline{74.9}   & {\cellcolor[rgb]{0.925,0.925,0.925}}\uline{70.5}  & {\cellcolor[rgb]{0.925,0.925,0.925}}\uline{77.6}  & {\cellcolor[rgb]{0.925,0.925,0.925}}\uline{32.2}  & {\cellcolor[rgb]{0.925,0.925,0.925}}\uline{41.9}   \\
                     & {\cellcolor[rgb]{0.925,0.925,0.925}}Ours [original]  & {\cellcolor[rgb]{0.925,0.925,0.925}}68.5          & {\cellcolor[rgb]{0.925,0.925,0.925}}74.9          & {\cellcolor[rgb]{0.925,0.925,0.925}}\textbf{70.7} & {\cellcolor[rgb]{0.925,0.925,0.925}}77.6          & {\cellcolor[rgb]{0.925,0.925,0.925}}31.7          & {\cellcolor[rgb]{0.925,0.925,0.925}}41.6           \\
                     & {\cellcolor[rgb]{0.925,0.925,0.925}}Ours [converted] & {\cellcolor[rgb]{0.925,0.925,0.925}}\textbf{69.1} & {\cellcolor[rgb]{0.925,0.925,0.925}}\textbf{75.5} & {\cellcolor[rgb]{0.925,0.925,0.925}}70.5          & {\cellcolor[rgb]{0.925,0.925,0.925}}\textbf{77.7} & {\cellcolor[rgb]{0.925,0.925,0.925}}\textbf{32.7} & {\cellcolor[rgb]{0.925,0.925,0.925}}\textbf{42.5}  \\
\bottomrule
\label{table:ZS-SBIR}
\end{tabular}
\\
\vspace{-0.2cm}
\scriptsize{*Indicates our reproduction, using codes provided by \cite{Tian_Xu_Shen_Yang_Shen_2022}.}
\end{table}

\noindent
\textbf{Categorical ZS-SBIR.}
Table \ref{table:ZS-SBIR} compares our approaches with SOTA methods in ZS-SBIR across diverse datasets.
The results indicate that the methods with CLIP (CLIP-AT~\cite{Sain_2023_CVPR} and ours) demonstrate superior performance, proving the effect of leveraging rich semantic information for enhancing zero-shot retrieval. Moreover, ours[clip] outperform all other methods, highlighting the efficacy of our proposed indirect alignment.
Notably, we observe a consistent improvement in mAP across all methods by a minimum of 8\% for Sketchy, 7\% for TU-Berlin, and 12\% for QuickDraw, proving our approach is more suitable than earlier triplet methods that directly correlate sketches and photos.

We also observe that our [original] and [converted] surpass [clip] version, indicating efficacy of the additional loss terms.
Furthermore, [converted] surpassing [original] implies our methods effectively address the modality gap.

\begin{table}
\vspace{-0.3cm}
\setlength{\defaultaddspace}{1.5pt}
\centering
\setlength{\extrarowheight}{0pt}
\addtolength{\extrarowheight}{\aboverulesep}
\addtolength{\extrarowheight}{\belowrulesep}
\setlength{\aboverulesep}{0pt}
\setlength{\belowrulesep}{0pt}
\scriptsize
\caption{Comparison for Generalized ZS-SBIR}
\label{table:GZS-SBIR}
\begin{tabular}{lcccc}
\toprule
Model &
\multicolumn{2}{c}
{Sketchy-Ext} & \multicolumn{2}{c}{TU-Berlin-Ext} \\ 
\cmidrule(lr){2-3}\cmidrule(lr){4-5}
& \multicolumn{1}{c}
{mAP@200}
& \multicolumn{1}{c}
{Prec@200}
& \multicolumn{1}{c}
{mAP@all}
& \multicolumn{1}{c}
{Prec@100} \\ 
\hline
SEM-PCYC~\cite{Dutta_2019_CVPR} & - & - & 19.2 & 29.8 \\
OCEAN~\cite{9102940} & - & - & 31.2 & 34.1 \\
BDA~\cite{chaudhuri22} & 22.6 & 33.7 & 25.1 & 35.7 \\
SaA~\cite{ribeiro2023sketchananchor} & - & - & 29.0 & 38.1 \\
ZSE[Ret]~\cite{Lin_2023_CVPR} & - & - & 46.4 & 48.5 \\
ZSE[RN]~\cite{Lin_2023_CVPR} & - & - & 43.2 & 46.0 \\
STL~\cite{Ge_Wang_Qi_Sun_Xu_Liao_2023} & \textbf{63.4} & 53.8 & 40.2 & 49.8 \\
CLIP-AT$^*$ & 55.6 & \uline{62.7} & \uline{60.9} & \uline{63.8} \\
\rowcolor[rgb]{0.925,0.925,0.925} Ours [converted] & \uline{62.3} & \textbf{68.5} & \textbf{62.6} & \textbf{67.8} \\
\addlinespace
\bottomrule
\end{tabular}
\vspace{-0.1cm}
\end{table}

\vspace{0.1cm} \noindent
\textbf{Generalized ZS-SBIR.}
In Table \ref{table:GZS-SBIR}, we observe that our method consistently outperforms baselines on the Generalized ZS-SBIR.
Results highlight again the validity of employing CLIP, evident when comparing the CLIP-AT~\cite{Sain_2023_CVPR} and ours with all other methods (excluding mAP@200 on Sketchy-Ext, where STL~\cite{Ge_Wang_Qi_Sun_Xu_Liao_2023} performs the best). Our method outperforms the previous state-of-the-art models by 12\% and 3\% in mAP, and by 9\% and 6\% in precision on Sketchy and TU, respectively, showing our method's suitability for both zero-shot learning and generalized settings.

\begin{table}
\centering
\setlength{\extrarowheight}{0pt}
\addtolength{\extrarowheight}{\aboverulesep}
\addtolength{\extrarowheight}{\belowrulesep}
\setlength{\aboverulesep}{0pt}
\setlength{\belowrulesep}{0pt}
\scriptsize
\caption{Comparison on the FG-ZS-SBIR task}
\begin{tabular}{lcc} 
\toprule
\multicolumn{1}{l}{Model} & \multicolumn{1}{c}{Acc@1} & \multicolumn{1}{c}{Acc@5} \\ 
\hline
CrossGrad~\cite{shankar2018generalizing} & 13.40 & 34.90 \\
CC-DG~\cite{Pang_2019_CVPR} & 22.60 & 49.00 \\
SketchPVT~\cite{Sain_2023_CVPR2} & \textbf{30.24} & 51.65 \\
CLIP-AT~\cite{Sain_2023_CVPR} & 28.68 & \textbf{62.34} \\
\rowcolor[rgb]{0.925,0.925,0.925} Ours [original] & \uline{29.96} & \uline{58.53} \\
\rowcolor[rgb]{0.925,0.925,0.925} Ours [converted] & 29.80 & 57.94 \\
\bottomrule
\end{tabular}
\label{table:FG-ZS-SBIR}
\vspace{-0.6cm}
\end{table}

\vspace{0.1cm} \noindent
\textbf{Fine-grained ZS-SBIR.}
In fine grained setting, we expect our method would not perform well, as our model is better suited for coarse-grained retrieval by design.
Surprisingly, however, Table~\ref{table:FG-ZS-SBIR} shows that ours comparably performs with state-of-the-art models specially designed for fine-grained setting, with minimal changes described in Sec.~\ref{sec:method:instance}.

Interestingly, we observe that our [original] embeddings outperform our [converted] in this setting, unlike the ZS-SBIR task.
We speculate that the conversion to the exact position in the target modality space is more challenging in the fine-grained setting, leading to a slight decline in performance with converted embeddings.

\begin{figure}
    \vspace{-0.1cm}
    \centering
    \includegraphics[width=0.85\linewidth]{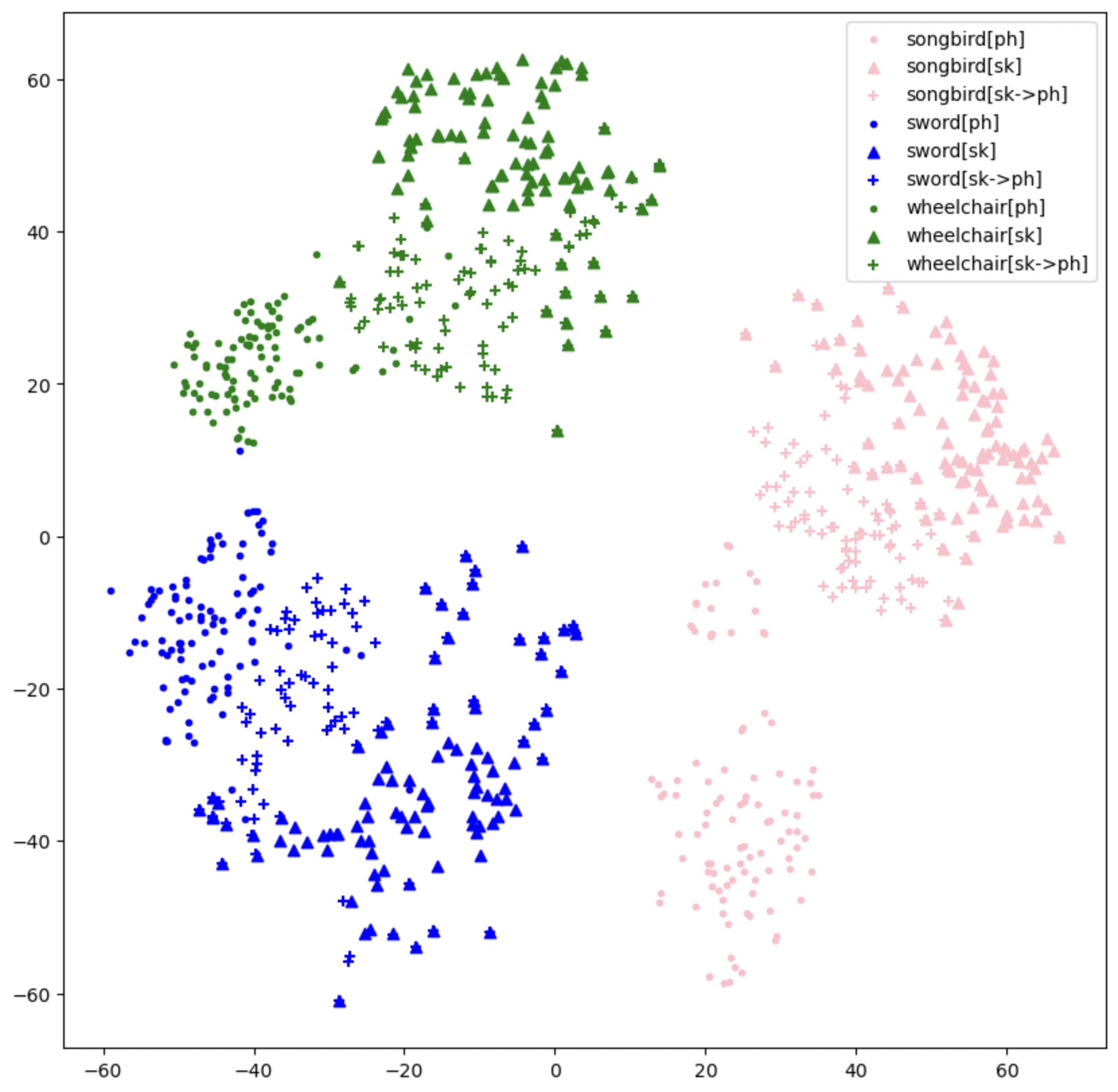}
    \caption{T-SNE visualization on Sketchy-Ext~\cite{Liu_2017_CVPR}}
    \label{fig:tsne}
    \vspace{-0.2cm}
\end{figure}

\vspace{0.1cm} \noindent
\textbf{Visualization.}
Fig.~\ref{fig:tsne} visualizes our learned embedding vectors corresponding to various modalities and classes.
We randomly select 100 vectors from \emph{songbird} (pink), \emph{sword} (blue), and \emph{wheelchair} (green) and plot with t-SNE projection.
Photos are marked with $\bullet$, sketches are with $\blacktriangle$, and the converted photo embeddings from sketches are with $+$, for each class.
We observe clear modality gap between photos and sketches, and also see that converted embeddings are clearly closer to the target, distinguished from the origin.

\subsection{Ablation Study}

\begin{table}
\centering
\scriptsize
\renewcommand{\tabcolsep}{2pt}
\caption{Ablation on Loss Terms for ZS-SBIR}
\vspace{0.1cm}

\begin{tabular}{ccccccccccc} 
\toprule
\multicolumn{5}{c}{\multirow{2}{*}{Loss}} & \multicolumn{2}{c}{Sketchy-Ext} & \multicolumn{2}{c}{TU-Berlin-Ext} & \multicolumn{2}{c}{QuickDraw-Ext} \\ 
\cline{6-11}
\addlinespace
$\mathcal{L}_\text{clip}$ & $\mathcal{L}_\text{sem}$ & $\mathcal{L}_\text{rec}$ & $\mathcal{L}_\text{mc}$ & $\mathcal{L}_\text{ortho}$ & \multicolumn{1}{c}{m@200} & \multicolumn{1}{c}{P@200} & \multicolumn{1}{c}{m@all} & \multicolumn{1}{c}{P@100} & \multicolumn{1}{c}{m@all} & \multicolumn{1}{c}{P@200} \\ 
\midrule
\checkmark & & & & & 68.50 & 74.90 & \textbf{70.53} & 77.62 & 32.17 & 41.85 \\
\checkmark & \checkmark & & & & 68.86 & 75.21 & \uline{70.52} & 77.67 & \uline{32.56} & \uline{42.44} \\
\checkmark & \checkmark & \checkmark & & & \textbf{69.05} & \uline{75.44} & 70.49 & \uline{77.68} & 32.20 & 42.29 \\
\checkmark & \checkmark & \checkmark & \checkmark & & \textbf{69.05} & \textbf{75.46} & 70.45 & 77.64 & 32.16 & 42.25 \\
\checkmark & \checkmark & \checkmark & \checkmark & \checkmark & \uline{69.02} & \uline{75.44} & 70.40 & \textbf{77.73} & \textbf{32.74} & \textbf{42.49} \\
\bottomrule
\end{tabular}
\label{table:ablation-loss}
\vspace{-0.3cm}
\end{table}

\vspace{0.1cm} \noindent
\textbf{Effect of Loss Terms.}
To evaluate the impact of individual loss terms, we conduct an ablation study to add loss terms one by one, reported in Table~\ref{table:ablation-loss}.
(Note that $\clip_l$ is always retained as an essential term for contrasting vectors.)

In general, a better result is achieved when more loss terms are used.
Taking a closer look, however, we observe slightly different patterns across datasets.
On Sketchy-Ext~\cite{Liu_2017_CVPR}, most proposed losses, except for the orthogonal loss, demonstrate effectiveness when added.
We speculate that the more stringent constraints imposed by the orthogonal loss slightly impede achieving superior outcomes.
On QuickDraw-Ext~\cite{Dey_2019_CVPR}, the inclusion of $\rec_l$ and $\disc_l$ leads to a slight decline in performance, while the orthogonal loss significantly improves the performance, unlike the case on Sketchy-Ext.
This suggests the potential interactions among these loss terms.
Experiments on TU-Berlin-Ext~\cite{Liu_2017_CVPR} show distinct trends.
Introducing more terms brings about a decline in mAP, while the opposite is observed in precision.

\begin{table}
\scriptsize
\centering
\caption{\centering ZS-SBIR on Sketchy-ext with unpaired datasets}
\begin{tabular}{lcc} 
\toprule
\multicolumn{1}{l}{Method} & mAP@200 & Prec@200 \\ 
\midrule
CLIP-AT$^{*}$ / 0.8 & 62.59 & 70.26 \\
Ours / 0.8 & 67.97 & 74.51 \\
Ours / 0.8 + photo / 0.2 & \uline{68.25} & \uline{74.67} \\
Ours / 0.8 + sketch / 0.2 & \textbf{68.56} & \textbf{75.11} \\
\bottomrule
\end{tabular}
\vspace{-0.3cm}
\label{table:ablation-ext}
\end{table}

\vspace{0.1cm} \noindent
\textbf{Effect of training on additional unpaired dataset.}
Our indirect alignment not only leads to improved performance but also facilitates semi-supervised approach, taking advantage of additional datasets containing only photographs or sketches.
To assess the impact of additional unpaired data, we conduct an experiment on Sketchy-Ext~\cite{Liu_2017_CVPR} as follows. First, a model is trained on 80\% of the seen categories (83 out of 104). Then, instances of either photos or sketches are added from the remaining seen classes, corresponding to the 20\% of the entire seen categories (21 out of 104).

As shown in Table \ref{table:ablation-ext}, seeing more samples even from unrelated classes improves the results. This is aligned with observations in previous studies~\cite{Bhunia_2021_CVPR, Sain_2023_CVPR, Sain_2023_CVPR2}.
We emphasize that other methods assuming paired sketches and photos are not eligible to take advantage of these unimodal datasets.

Taking a deeper look, adding more sketches yields greater advantage than adding more photos.
Augmenting sketches is more efficient but expensive. We leave further investigation as a potential future work.

\begin{table}
\centering
\scriptsize
\renewcommand{\tabcolsep}{3pt}
\caption{Ablation on Model Types for ZS-SBIR}
\begin{tabular}{lcccccc} 
\toprule
\multicolumn{1}{c}{\multirow{2}{*}{Loss}} & \multicolumn{2}{c}{Sketchy-Ext\cite{Liu_2017_CVPR}} & \multicolumn{2}{c}{TU-Berlin-Ext\cite{Liu_2017_CVPR}} & \multicolumn{2}{c}{QuickDraw-Ext\cite{Dey_2019_CVPR}} \\ 
\cline{2-7}
\addlinespace
\multicolumn{1}{c}{} & \multicolumn{1}{c}{m@200} & \multicolumn{1}{c}{P@200} & \multicolumn{1}{c}{m@all} & \multicolumn{1}{c}{P@100} & \multicolumn{1}{c}{m@all} & \multicolumn{1}{c}{P@200} \\ 
\midrule
Ours full & \textbf{69.05} & \textbf{75.46} & \textbf{70.49} & \textbf{77.68} & \textbf{32.74} & \textbf{42.49} \\
Ours ($g$ frozen) & 67.25 & 74.23 & 68.72 & 75.12 & 30.83 & 37.43 \\
\bottomrule
\end{tabular}
\label{table:ablation-txt-sbir}
\vspace{-0.2cm}
\end{table}

\begin{table}
\centering
\scriptsize
\setlength{\extrarowheight}{0pt}
\addtolength{\extrarowheight}{\aboverulesep}
\addtolength{\extrarowheight}{\belowrulesep}
\setlength{\aboverulesep}{0pt}
\setlength{\belowrulesep}{0pt}
\caption{Ablation on Textual Intervention}
\vspace{0.1cm}
\begin{tabular}{cccc} 
\toprule
$\lambda_{\text{clip-img}}$ & 1.0 & 0.8 & 0.5 \\
\midrule
Acc@1 & 29.96 & 27.31 & 24.71 \\
Acc@5 & 58.53 & 54.79 & 70.71 \\
\bottomrule
\end{tabular}
\label{table:ablation-txt-fg-sbir}
\vspace{-0.4cm}
\end{table}

\vspace{0.1cm} \noindent
\textbf{Effect of textual supervision.} 
In categorical SBIR, our proposed model aligns sketches and photos solely via texts. We hypothesize that the text encoder would play an important role, since the image distribution should, to some extent, follow the fixed text distribution if the text encoder is frozen and the text distribution is fixed.


Table \ref{table:ablation-txt-sbir} evaluates the effect of fine-tuning the text encoder.
As expected, the performance slightly drops when the text encoder remains frozen.
We interpret that realigning the text encoder leads to a more effective utilization of latent space, perhaps achieving higher uniformity~\cite{pmlr-v119-wang20k} and better adaptation to SBIR datasets, while simultaneously preventing catastrophic forgetting.

For FG-SBIR, we explore several different values for $\lambda_\text{clip-img}$ in Table~\ref{table:ablation-txt-fg-sbir}.
We observe that alignment with the class hinders the exact alignment of identical instances between sketches and photos, even when we unfreeze the text encoder.
This is in contrast to \cite{Sain_2023_CVPR}, which suggests that such alignment facilitates accurate alignment in the latent space.

\subsection{Qualitative Results}
\begin{figure}
    \vspace{-0.3cm}
    \includegraphics[width=\linewidth]{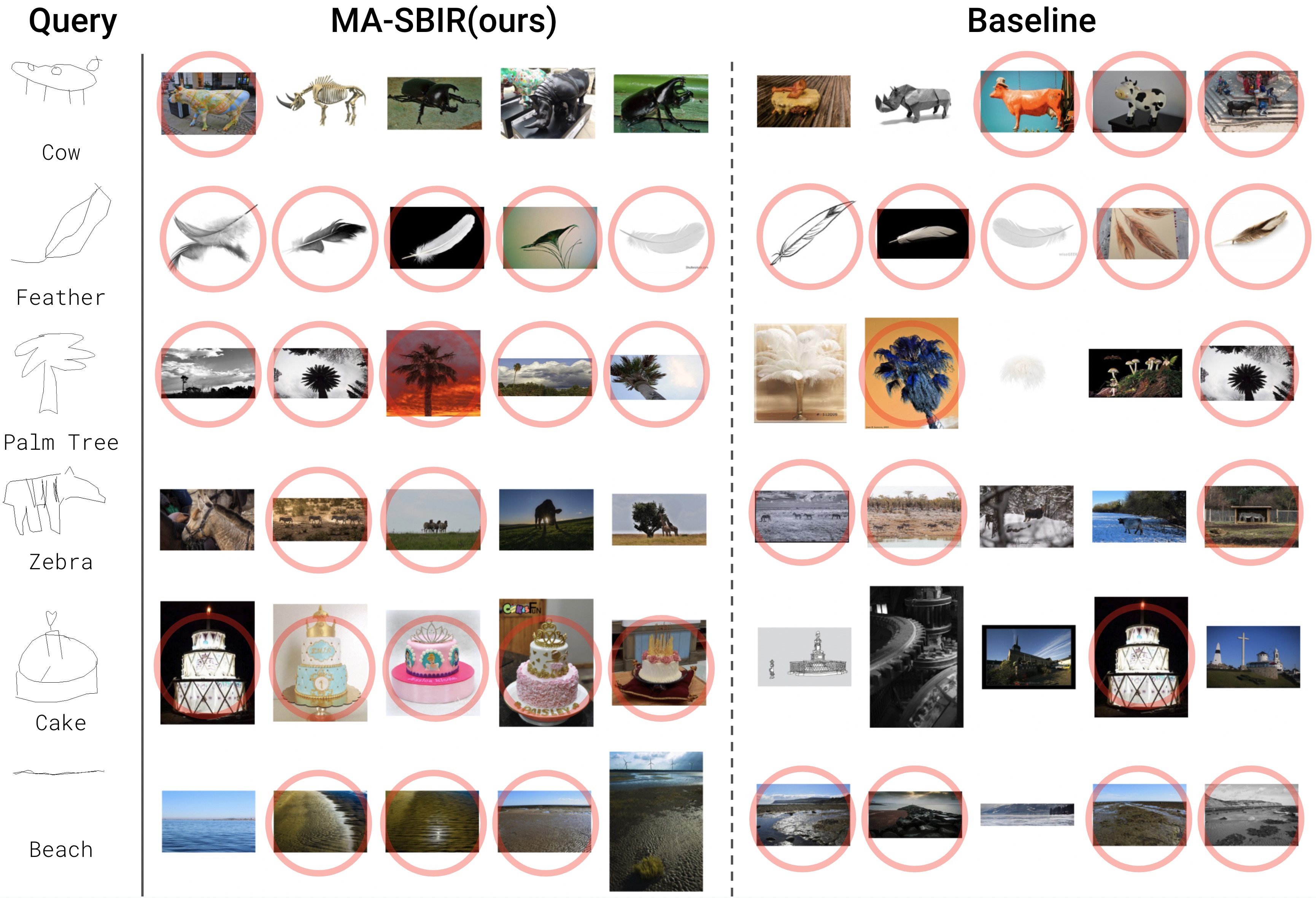}
    \caption{
        \textbf{Categorical-ZS-SBIR.} Top-5 Retrieved images on QuickDraw-Ext~\cite{Dey_2019_CVPR}. Correct samples are circled.
    }
    \label{fig:qual-sbir-mini}
    \vspace{-0.3cm}
\end{figure}
\begin{figure}
    \includegraphics[width=\linewidth]{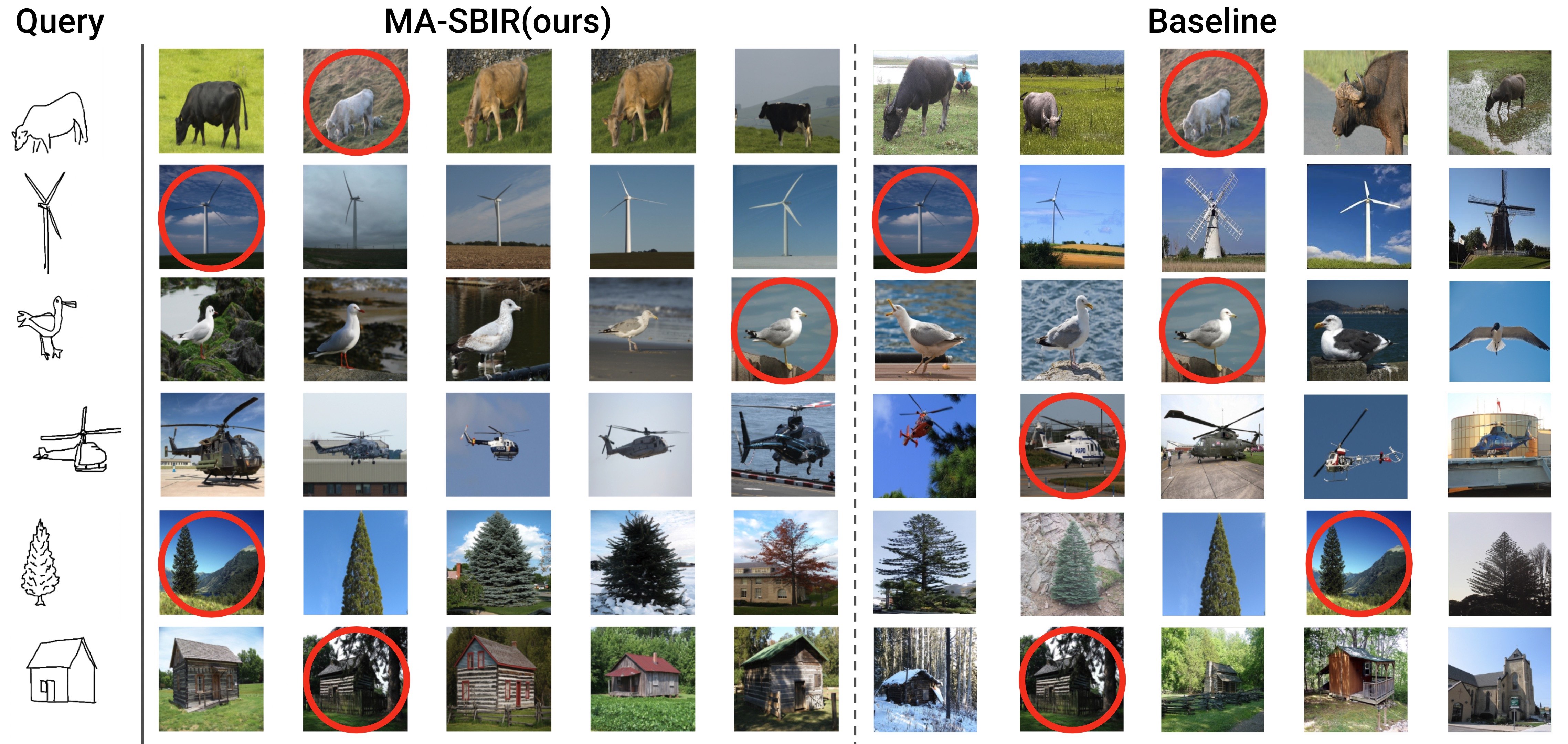}
    \caption{
        \textbf{FG-ZS-SBIR.} Top-5 Retrieved images on Sketchy~\cite{Dey_2019_CVPR}. Correct samples are circled.
    }
    \label{fig:qual-fg-sbir-mini}
    \vspace{-0.5cm}
\end{figure}

We qualitatively compare our retrieval results with the strongest baseline, CLIP-AT~\cite{Sain_2023_CVPR}, in Fig.~\ref{fig:qual-sbir-mini} (categorical) and Fig.~\ref{fig:qual-fg-sbir-mini} (fine-grained).
Refer to the supplementary material for more examples. Overall, our method retrieves correct photos for most classes.

For the fine-grained task, even incorrectly marked images turn out to be often correct.
For instance, for a drawing of a wind turbine (row 2 in Fig.~\ref{fig:qual-fg-sbir-mini}), all 5 photos retrieved by our model are indeed wind turbines, although they are not labeled in the ground truth. The baseline also retrieves 3 wind turbine images, but it includes two windmill images as well (at the 3rd and 5th).
In row 3 and 4 of Fig.~\ref{fig:qual-fg-sbir-mini}, our method ranks the true image lower than the baseline does, but the retrieved photos actually exhibits greater visual similarity with the queried sketches.
\vspace{-0.2cm}
\section{Summary and Limitations}
\label{sec:conclusion}

\vspace{-0.1cm}

We present a novel approach that aligns the joint embedding space by disentangling modality-specific nuances from semantic content.
Our method excels in multiple zero-shot scenarios of sketch-based image retrieval, setting a new performance benchmark.

Despite the success on categorical setting, however, our model turns out not to align well at instance level.
This is somewhat expected, as our model is designed to be trainable with unpaired sketches and photos, but this is not applicable on fine-grained setting.
Improving fine-grained alignment without paired examples will be an interesting future work.


\vspace{0.1cm}
{ \footnotesize
\noindent\textbf{Acknowledgement.} This work was supported by the New Faculty Startup Fund from Seoul National University and by National Research Foundation (NRF) grant (No. 2021H1D3A2A03038607/50\%, 2022R1C1C1010627/20\%, RS-2023-00222663/10\%) and Institute of Information \& communications Technology Planning \& Evaluation (IITP) grant (No. 2022-0-00264/20\%) funded by the government of Korea.}

{\small
\bibliographystyle{ieee_fullname}
\bibliography{ref}
}

\end{document}